\documentclass[singlecolumn]{article}

\usepackage[english]{babel}

\usepackage[letterpaper,top=2cm,bottom=2cm,left=3cm,right=3cm,marginparwidth=1.75cm]{geometry}

\usepackage{amsmath}

\usepackage{graphicx}
\usepackage{xcolor}
\usepackage[colorlinks=true, allcolors=blue]{hyperref}
\usepackage{amssymb}
\usepackage{multirow}
\usepackage{subfig}
\usepackage{stfloats}
\usepackage{url}
\usepackage{authblk}

\title{Adaptive Output Steps: FlexiSteps Network for Dynamic Trajectory Prediction}
\author[1]{Yunxiang Liu}
\author[2]{Hongkuo Niu \thanks{Corresponding author: 236142132@.mail.sit.edu.cn}}
\author[3]{Jianlin Zhu}

\affil[1]{Faculty of Intelligence Technology, Shanghai Institute of Technology}
\affil[2]{Faculty of Intelligence Technology, Shanghai Institute of Technology}
\affil[3]{Faculty of Intelligence Technology, Shanghai Institute of Technology}

\begin{document}
\maketitle

\begin{abstract}
Accurate trajectory prediction is vital for autonomous driving, robotics, and intelligent decision-making systems, yet traditional models typically rely on fixed-length output predictions, limiting their adaptability to dynamic real-world scenarios. In this paper, we introduce the FlexiSteps Network (FSN), a novel framework that dynamically adjusts prediction output time steps based on varying contextual conditions. Inspired by recent advancements addressing observation length discrepancies and dynamic feature extraction, FSN incorporates an pre-trained Adaptive Prediction Module (APM) to evaluate and adjust the output steps dynamically, ensuring optimal prediction accuracy and efficiency. To guarantee the plug-and-play of our FSN, we also design a Dynamic Decoder(DD). Additionally, to balance the prediction time steps and prediction accuracy, we design a scoring mechanism, which not only introduces the Fréchet distance to evaluate the geometric similarity between the predicted trajectories and the ground truth trajectories but the length of predicted steps is also considered. Extensive experiments conducted on benchmark datasets including Argoverse and INTERACTION demonstrate the effectiveness and flexibility of our proposed FSN framework. 
\end{abstract}

\section{Introduction}

Trajectory prediction plays an essential role in various critical applications such as autonomous driving, robotics, and intelligent decision-making systems. Accurately predicting the future motion of dynamic agents is fundamental to ensuring safety and efficiency in real-world scenarios. Recent advancements in deep learning have significantly improved the precision of trajectory prediction models.\cite{chai2019multipath,varadarajan2022multipath++,zhou2022hivt,zhou2023query,zhou2024smartrefine,tang2024hpnet,wang2025dynamics,climent2025learning,liu2025gamdtp,zhu2025dyttp} However, most existing methods are constrained by a fixed-length prediction horizon, limiting their adaptability and effectiveness when confronted with dynamic and unpredictable environments.

\begin{figure*}[t]
\centering
\includegraphics[width=\textwidth]{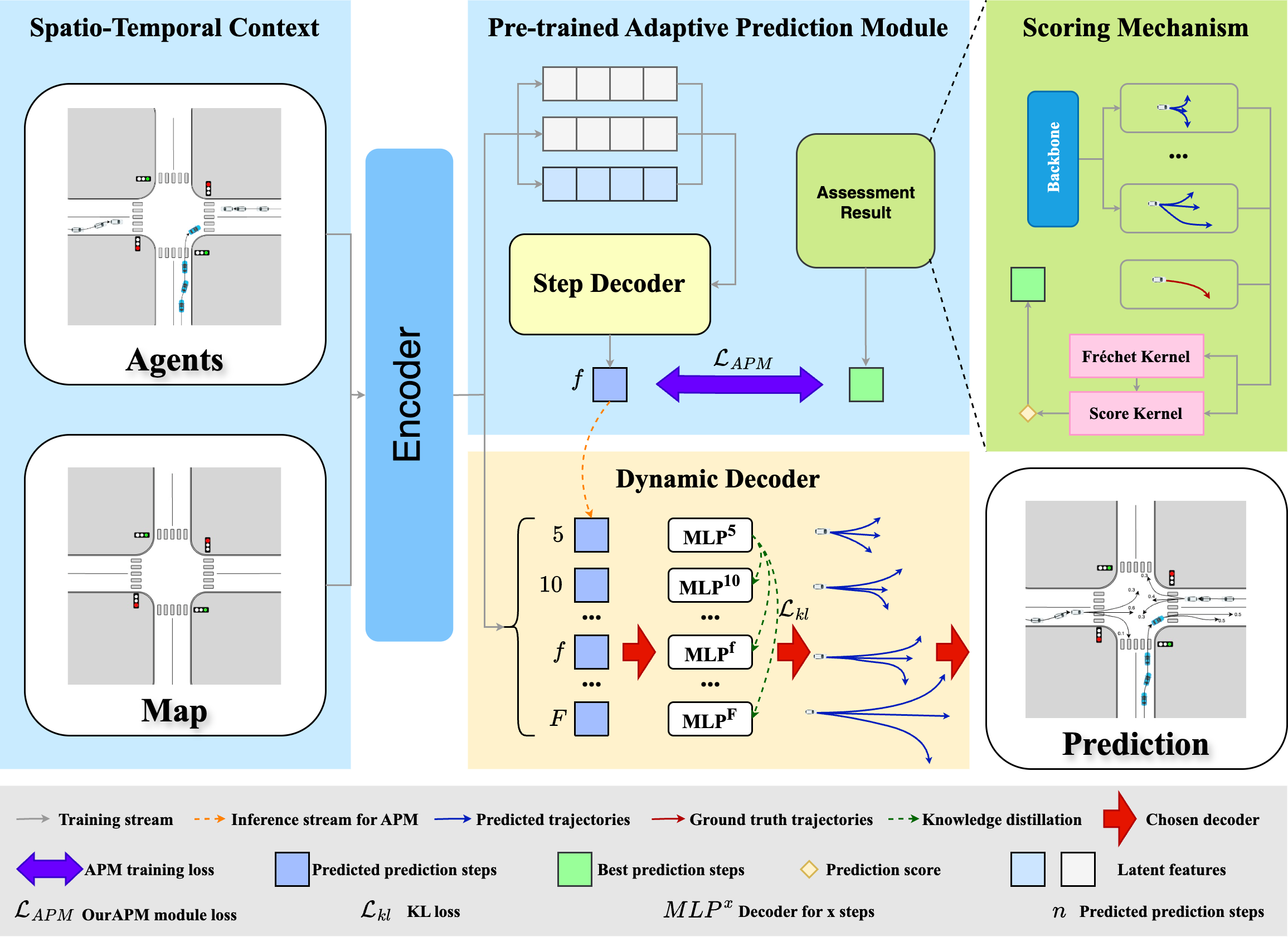}
\caption{\label{FSN}The overview of our FlexiSteps Network(FSN).}
\end{figure*}

Traditional trajectory prediction models typically assume a consistent length for output predictions, neglecting variations that naturally occur due to changing contextual conditions. This limitation can result in insufficient or excessive use of computational resources, thereby reducing practical applicability. Recently, studies have begun addressing discrepancies in observation lengths (input data), revealing significant performance degradation when the models are tested under conditions differing from those during training, issue termed "Observation Length Shift". To address this, FlexiLength Network (FLN)\cite{xu2024adapting} introduced calibration and adaptation mechanisms to accommodate varying observation lengths, as shown in Fig.~\ref{FSN}, showing promising improvements in prediction robustness and flexibility.

\begin{figure}[t]
\centering
\subfloat[HiVT]{\includegraphics[width=.45\columnwidth]{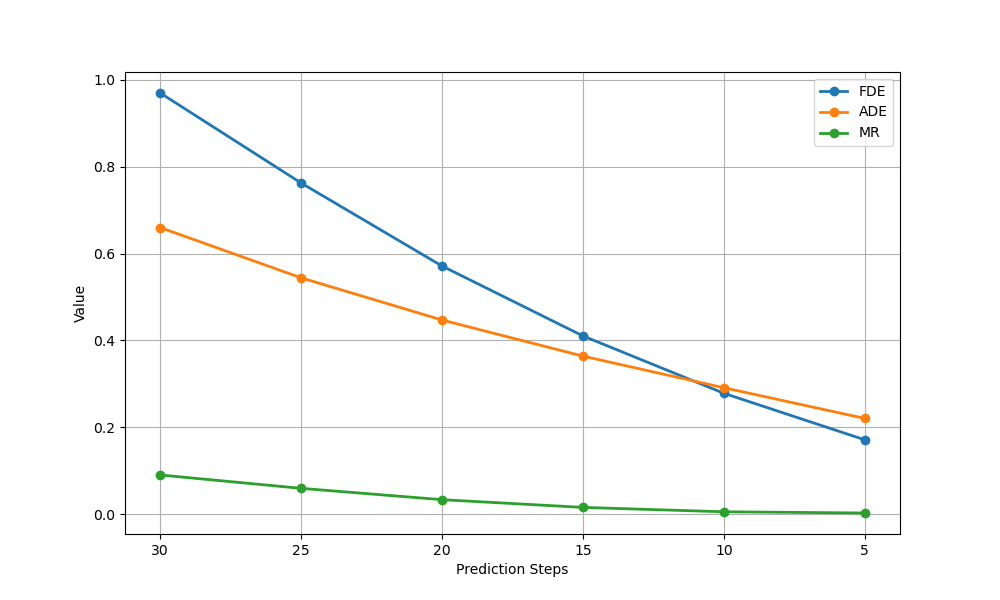}}\hspace{5pt}
\subfloat[HPNet]{\includegraphics[width=.45\columnwidth]{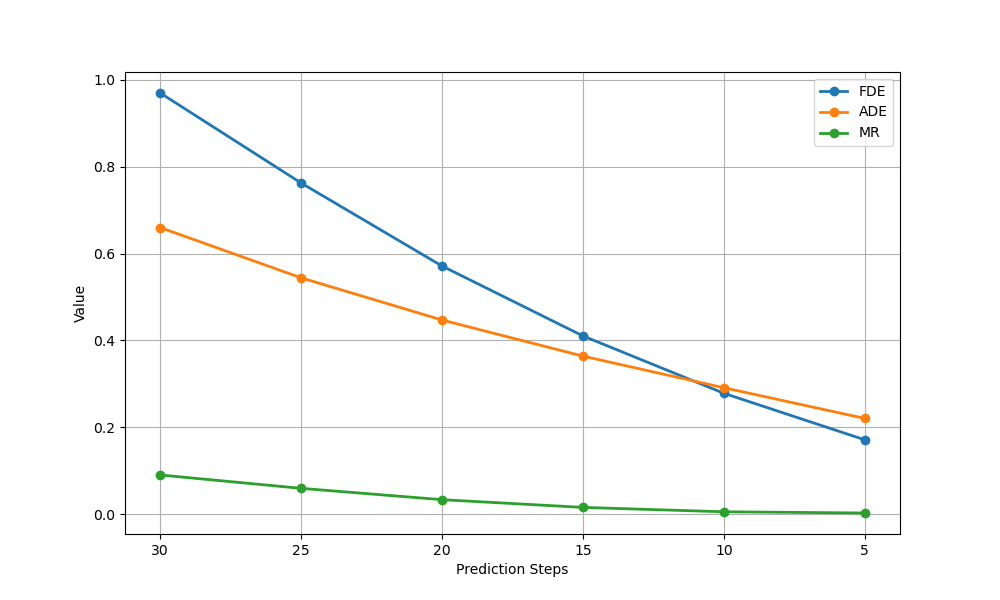}}
\caption{Prediction results of HiVT and HPNet with different fixed prediction steps.}
\label{result_diff_step}
\end{figure}

Similarly, Length-agnostic Knowledge Distillation (LaKD)\cite{li2024lakd} was proposed to handle varying observation lengths by dynamically transferring knowledge between trajectories of differing lengths, thus overcoming the inherent limitations of traditional fixed-length input methods. LaKD's approach highlighted that the effectiveness of longer observed trajectories could sometimes be compromised by interference, emphasizing the necessity for adaptive mechanisms to handle real-world trajectory variations.

Inspired by these developments, we propose the FlexiSteps Network (FSN), a novel trajectory prediction framework specifically designed to dynamically adjust the prediction output steps based on contextual cues and environmental conditions. FSN incorporates an innovative pre-trained Adaptive Prediction Module (APM) to evaluate and dynamically modify the number of predicted future steps during inference, ensuring predictions remain optimal in terms of accuracy and computational efficiency. Further, to guarantee the plug-and-play of our FSN, we also design a Dynamic Decoder(DD). During the training phase, the decoder is trained to output sequences with varying step lengths, thereby facilitating dynamic output generation during inference.

Howerver, the challenge of balancing prediction horizon and accuracy remains a critical issue. Traditional metrics such as Average Displacement Error (ADE) and Final Displacement Error (FDE) often fail to capture the overall shape and temporal consistency of predicted trajectories, leading to suboptimal evaluation of model performance\cite{he2024frechet}. Recent researches\cite{bhattacharyya2018lon,kuo2022trajectory} and our experiments, as shown in Fig.~\ref{result_diff_step}, also show that prediction accuracy and prediction horizon are inversely proportional. To address this, we introduce a scoring mechanism that leverages the Fréchet distance, a robust geometric measure that comprehensively assesses trajectory similarity by considering both spatial and temporal relationships.\cite{eiter1994computing,10939571} By combining the Fréchet distance with the prediction steps, we can effectively evaluate the quality of predictions while dynamically adjusting the output horizon.

This scoring mechanism not only enhances the evaluation of trajectory predictions but also provides a means to trade off between prediction horizon and accuracy. By incorporating the Fréchet distance, we can ensure that the model does not favor shorter-term predictions for higher precision, thus maintaining a balance between flexibility and accuracy.

Through extensive experiments on prominent benchmark datasets, including Argoverse and INTERACTION, our FSN demonstrates significant improvements in flexibility and predictive accuracy compared to traditional models. This framework provides a practical solution to the critical need for adaptive, context-aware trajectory prediction models, setting a new benchmark for effectiveness in dynamic prediction scenarios.

In summary, our work has the following contributions:
\begin{itemize}
\item We propose a novel dynamic trajectory prediction framework FlexiSteps Network(FSN), which includes pre-trained Adaptive Prediction Module(APM) and Dynamic Decoder(DD) and both are plug-and-play for learning-based models.
\item We also introduce a scoring mechanism to trade off the prediction horizon and accuracy. Due to the fréchet distance can consider the spatial and temporal relationships, we combine it and the prediction steps to implement the socring mechanism.
\item We validate the accuracy and flexibility through comprehensive experiments on benchmark datasets including Argoverse\cite{chang2019argoverse} and INTERACTION\cite{zhan2019interaction}.
\end{itemize}

\section{Related Work}
\subsection{Traditional Trajectory Prediction}
Trajectory prediction plays a critical role in applications such as autonomous driving, robotics, and intelligent systems. Traditionally, trajectory prediction approaches focus on using deep neural networks to learn from historical agent movements and contextual data, effectively modeling complex interactions and behaviors. Methods leveraging Graph Neural Networks (GNNs)\cite{liang2020learning,wang2020multi,xu2023mvhgn,zhang2022ai,sheng2022graph}, Generative Adversarial Networks (GANs)\cite{goodfellow2014generative,gupta2018social}, and Transformer-based architectures have significantly advanced the field\cite{azadani2023stag,gu2021densetnt,zhou2022hivt,ngiam2021scene,mohamed2020social,gao2023dual,zhou2022spatiotemporal,hu2022trajectory,hou2022structural}, demonstrating robust predictive capabilities on various benchmarks.
\subsection{Variable Timestep Prediction}
A major limitation of traditional trajectory prediction methods is their inability to handle varying data effectively. Xu and Fu\cite{xu2024adapting} propose the FlexiLength Network (FLN), which integrates trajectory data of diverse length and employs FlexiLength Calibration (FLC) and Adaptation (FLA) to learn temporal invariant representations. In contrast, Li et al.\cite{li2024lakd} introduce LaKD, a length-agnostic knowledge distillation framework that dynamically transfers knowledge between trajectories of different lengths. LaKD addresses the limitations of FLN by employing a length-agnostic knowledge distillation to dynamically transfer knowledge and a dynamic soft-masking mechanism to prevent knowledge conflicts. However, both FLN and LaKD primarily focus on input length variability and do not address the challenge of dynamically adjusting output prediction steps, which is critical for real-time decision-making.\par
Other similar methods such as \cite{karle2024self} also try to select different prediction models by understanding different scenarios, but they can not achieve plug-and-play effects for more learning-based methods, which have great limitations. And \cite{wang2025dynamics} just divided the prediction into 2 key horizons including short-term and long-term, which is not flexible enough to adapt to the real-world scenarios. In contrast, our proposed FlexiSteps Network (FSN) introduces a pre-trained Adaptive Prediction Module (APM) that dynamically adjusts the output prediction steps based on contextual cues, ensuring optimal prediction accuracy and efficiency. This approach allows for greater flexibility in handling varying prediction requirements, making FSN a more adaptable solution for dynamic trajectory prediction tasks.
\subsection{Metrics for Trajectory Prediction}
Traditional metrics such as ADE, FDE, MR, has been used in almost all trajectory prediction methods\cite{qiao2014self,jiang2022vehicle,zyner2019naturalistic,chen2022intention,zhang2022trajectory,wang2020multi,azadani2023stag,zhao2021tnt,gu2021densetnt} in autonomous driving scenarios. However, these metrics overlook the overall shape and temporal consistency\cite{he2024frechet}. Song et al.\cite{song2025don} employs Hausdorff Distance\cite{huttenlocher1993comparing,dubuisson1994modified,belogay1997calculating} in their trajectory matching module to ensure coherence. While the Hausdorff distance serves as a natural metric for computing curves or compact sets, it ignores both the direction and motion dynamics along the curves\cite{shahbaz2013applied,zhao20183d}.\par
The Fréchet distance, however, accounts for the order of points, making it a superior measure for assessing curve similarity\cite{shahbaz2013applied}. Thus it is a robust geometric measure, has been proposed as an alternative to ADE and FDE, providing a more comprehensive evaluation of trajectory prediction quality\cite{bhattacharyya2018long,kuo2022trajectory,alt1995computing}. By considering both spatial and temporal relationships, the Fréchet distance offers a more nuanced assessment of trajectory predictions, making it a suitable choice for evaluating dynamic prediction models.

\section{Method}

Our FlexiSteps Network (FSN) is designed to dynamically adjust the prediction output steps based on contextual cues and environmental conditions. The overall architecture of FSN is shown in Fig.~\ref{FSN} and Sec.\ref{method:overview}. The key components of FSN include a pre-trained Adaptive Prediction Module (APM), a Dynamic Decoder (DD), and a scoring mechanism that incorporates the Fréchet distance to evaluate trajectory predictions. We detail scoring mechanism in Sec.\ref{method:dd}, APM in Sec.\ref{method:apm} and DD in Sec.\ref{method:scoring} respectively.
\subsection{Problem Formulation}
Given the target agent $i$'s locations $[l_i^0,..,l_i^{T-1},l_i^T]$ in the past $T$ time steps, where $l_i^t \in \mathbb{R}^2$ represents the 2D coordinates of the agent at time step $t$, we aim to predict the future locations $[l_i^{T+1},...,l_i^{T+F}]$, where $F$ is the number of future steps to be predicted. We present the historical relative trajectory of agent $i$ as $p_i=\left\{ l_i^t - l_i^{t-1} \right\}_{t=1}^{T}$ in historical time steps. Naturally, the target agent will interact with the context including historical locations of surrounding agents and high-definition map(HD map) that represented as $p_{oth} = [p_0,...p_{N_a}]$ and $p_{\xi}=\left\{ p^1_\xi - p^0_\xi\right\}_{\xi=0}^{N_m}$, where $N_a$ is the number of surrounding agents within a certain radius, $N_m$ is the number of HD map segments in the same radius, and $p^0_\xi$ and $p^1_\xi$ are the start and end points of the HD map segment $\xi$.
\subsection{Overview of FlexiSteps Network}\label{method:overview}
The overview of our FlexiSteps Network (FSN) is shown in Fig.~\ref{FSN}. Most methods in trajectory prediction use the framework of encoder-decoder, where the encoder encodes the historical trajectory and context information into a latent representation and the decoder decodes it into future trajectory. Our FSN also follows this way, and we focus on the decoder method, which means that we use baseline encoder as:
\begin{equation}
    e_{i,k} = \phi_{enc}([P_i, c_i], [p_{\xi}, c_{\xi}]),
\end{equation}
where $k$ is the mode index in multi-modal prediction framework, $e_{i,k}$ is the latent representation of the target agent $i$ under the $k$ mode, $\phi_{enc}$ is the encoder function from our baseline methods, including\cite{zhou2022hivt,tang2024hpnet}, $P_i$ is concated from $p_i$ and $p_{oth}$, $c_i$ is the agent attribute including agent heading, velocity and agent type, and $c_{\xi}$ is the HD map lane segments attribute including lane heading, lane turn direction and whether it is an interaction. The output of the encoder is then fed into our APM and DD module to generate prediction steps and future trajectory predictions.\par
The FSN framework consists of three main components: the pre-trained Adaptive Prediction Module (APM), the Dynamic Decoder (DD), and the scoring mechanism. The APM is responsible for dynamically adjusting the prediction output steps based on contextual cues and environmental conditions. It is trained through the assessment result from trained baseline model in different fixed output steps. And the assessment result is evaluated by the scoring mechanism. The DD is designed to handle varying prediction lengths, allowing the model to output sequences with different step lengths during inference. This flexibility is crucial for adapting to real-world scenarios where the required prediction horizon may vary significantly. Finally, the scoring mechanism incorporates the Fréchet distance to evaluate the quality of trajectory predictions, considering both spatial and temporal relationships. By combining these components, FSN provides a robust and adaptable solution for dynamic trajectory prediction tasks.

\begin{figure}[t]
    \centering
    \includegraphics[width=0.5\textwidth]{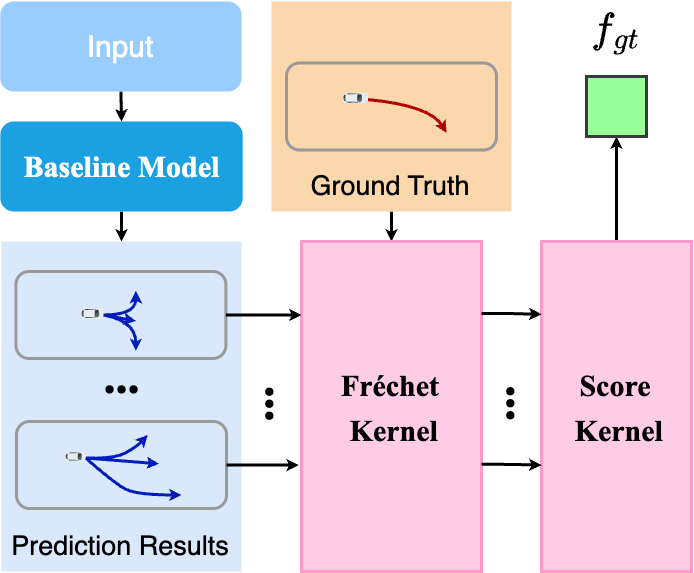}
    \caption{Overview of our scoring mechanism. The Fréchet distance is used to evaluate the quality of trajectory predictions, and the prediction length is also considered to balance accuracy and prediction horizon.}
    \label{score_flow}
\end{figure}

\subsection{Scoring Mechanism}\label{method:scoring}
According to the experiments on HiVT\cite{zhou2022hivt} and HPNet\cite{tang2024hpnet}, as shown in Fig.~\ref{result_diff_step} the prediction accuracy and prediction horizon are inversely proportional, which means that the longer the prediction horizon, the lower the prediction accuracy. This will cause our Dynamic Decoder to tend to predict shorter durations for lower prediction errors and influence the performance of our Dynamic Prediction. To address this, we introduce a scoring mechanism that incorporates the Fréchet distance to evaluate the quality of trajectory predictions. As shown in Fig.~\ref{score_flow}\par
\subsubsection{Fréchet Distance Kernal}
The Fréchet distance is a robust geometric measure that comprehensively assesses trajectory similarity by considering both spatial and temporal relationships\cite{eiter1994computing,10939571}. Howerver, it can not be directly applied into machine learning frameworks due to its non-smoothness function as:
\begin{equation}
    d_F(X, Y) := \mathop{min}\limits_{A \in \mathcal{A}}\mathop{max}\limits_{(i,j) \in A} d(x_i, y_j)
\end{equation} 
where $d(x,y)$ is the Euclidean distance between vector $x$ and $y$, $X = \{x_1, x_2, ..., x_m\}$ and $Y = \{y_1, y_2, ..., y_n\}$ are two sets of points, and $\mathcal{A}$ is the set of all possible alignments between the two sets. Fréchet distance kernal (FRK) \cite{10.1145/3474717.3483949} design soft-min approximation and smooth-min approximation to make fréchet distance computable in machine learning frameworks. Howereve, the smooth-min approximation of the original FRK is noise-sensitive because the exponential weight assigned to outliers may be too large. To address this issue, we propose a new Fréchet distance kernal (FDK) to approximate the fréchet distance as a smooth function by introducing Huber Loss Smooth \cite{gupta2020robust,meyer2021alternative}:

\begin{equation}
    H(z,\delta)=
    \begin{cases}
    \frac{1}{2}z^2 & \text{if } |z| \leq \delta \\
    \delta(|z|-\frac{1}{2}\delta) & \text{otherwise}
    \end{cases}
\end{equation}
where $\delta$ is threshold parameters, which is set to 0.1 in our experiments. The Fréchet distance kernal (FDK) is defined as:
\begin{equation}
    \resizebox{.85\hsize}{!}{
    $\mathop{min}\limits_{(i,j)\in A}\phi(x_i,y_j) \approx \frac{\sum\limits_{(i,j\in A)}\phi(x_i,y_j) \cdot exp(-\beta \cdot H(\phi(x_i,y_j),\delta))}{Z_A}$}
\end{equation}
\begin{equation}
    \resizebox{.85\hsize}{!}{
    $FDK(X,Y) := \sum\limits_{A \in \mathcal{A}} \sum\limits_{(i,j) \ in A} \frac{\widetilde{\phi}_\epsilon(x_i,y_j) exp(-\beta \cdot H(\phi(x_i,y_j),\delta))}{Z_A}$}
\end{equation}
where $\phi(x_i,y_j) = exp(-d(x_i,y_j)/\gamma)$, $\widetilde{\phi}_\epsilon(x_i,y_j) := exp(-d(x_i,y_j)/\gamma+\epsilon\delta(x_i - y_j))$ with parameters $\epsilon > 0$, $\beta$, $Z_A > 0$.

\subsubsection{Score Kernal}
In addition, in order to balance accuracy and prediction length, it is not enough to just calculate the fréchet distance, the prediction length is necessary to be included. The smaller the distance, the closer between the prediction trajectories and ground truth trajectories. To make the fraction of the prediction trajectory smaller, the higher the prediction quality, we supplement the prediction length at the denominator as:
\begin{equation}
    q_i^f = \frac{d_i^f}{f}
\end{equation}
where $q_i^f$ is the score of agent $i$ with prediction steps $f$, $d_i^f = FDK(\mu_i^f, gt_i^f)$ is the Fréchet distance between the predicted trajectory $\mu_i^f$ and the ground truth trajectory $gt_i^f$ of agent $i$ with prediction steps $f$. The score $q_i^f$ is designed to be lower for better predictions, as it combines the Fréchet distance with the prediction length, thus penalizing longer predictions that do not match the ground truth well.\par
Finally, we can get the final score from all prediction steps:
\begin{equation}
    q_i = min(q_i^f|_{f=5}^F)
\end{equation}
And the prediction step of agent $i$ is:
\begin{equation}
    {f_{gt}}_i = \operatorname{argmin}_{f \in \{5,...,F\}} q_i^f
\end{equation}
To summarize, we express the scoring mechanism in:
\begin{equation}
    {f_{gt}}_{i} = \phi_{score}(\mu_{i,k}^f, gt_i^f)
\end{equation}
where ${f_{gt}}_{i}$ is the optimal prediction step achieves the lowest score, i.e.the best result, under our scoring mechanism, $\phi_{score}$ is the scoring mechanism function, and $gt_i^f = \left\{l_i^t\right\}_{t=T+1}^{T+f}$ is the ground truth future trajectory of agent $i$. 
By combining the Fréchet distance with the prediction steps, we can effectively evaluate the quality of predictions while dynamically adjusting the output horizon.\par

\subsection{Pretrained Adaptive Prediction Module}\label{method:apm}
The Adaptive Prediction Module (APM) is a pre-trained module that adaptively adjusts the prediction steps based on the contextual information and environmental conditions. The APM is trained using a set of historical trajectories and their corresponding future trajectories, allowing it to learn the optimal prediction steps for different scenarios. The training process involves evaluating the performance of the baseline model with different fixed prediction steps and using the assessment results to guide the APM's adjustments. And the detailed evaluation criteria will illustrated in Sec.\ref{method:scoring}.\par

\subsubsection{Training stage}
To train our APM, we first collect the prediction results from the baseline model trained by different fixed output steps:
\begin{equation}
    \mu_{i,k}^f, b_{i,k}^f = \phi_{D^f}(\phi_{enc}([P_i, ci], [p_{\xi}, c_{\xi}]))    \label{eq:fixed_output}
\end{equation}
where $\phi_{D^f}$ is the decoder function from our baseline methods, $f \in [1,F]$ is the fixed output steps, $\mu_{i,k}^f$ and $b_{i,k}^f$ are the predicted future trajectory and probability of agent $i$ mode $k$ with fixed output steps $f$ respectively.\par

\begin{figure}[t]
    \centering
    \includegraphics[width=0.5\textwidth]{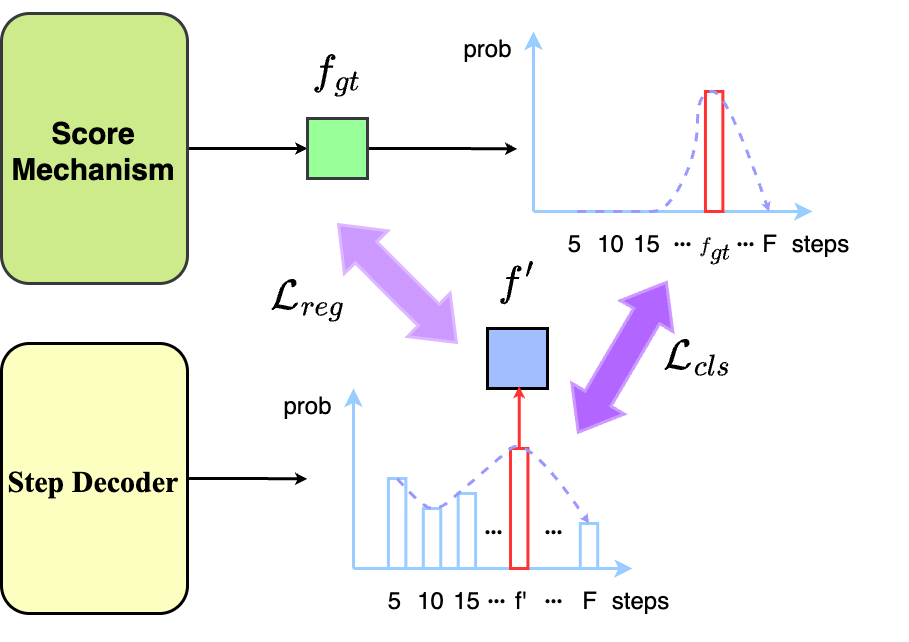}
    \caption{Training stage of the Adaptive Prediction Module (APM). The APM is trained to predict the optimal prediction step based on the encoded latent features of the target agent and its context.}
    \label{training_stage}
\end{figure}

Then we use the scoring mechanism to evaluate the prediction results and return the optimal prediction steps for each agent ${f_{gt}}_i$. After collecting the optimal prediction steps for all agents among datasets, we can train our APM as illustrated in Fig.~\ref{training_stage} and following:
\begin{equation}
    b'_i = \phi_{step}(e_{i,k})
\end{equation}
\begin{equation}
f'_i = \operatorname{argmax}_{j \in \{1,2,\ldots,F\}} {b'_i}^j
\end{equation}
\begin{equation}
    \mathcal{L}_{cls} = -\frac{1}{N}\sum_{i=1}^{N}\sum_{j=1}^{F} b_i^j \log({b'_i}^j)
    \end{equation}
\begin{equation}
    \mathcal{L}_{reg} = \frac{1}{N}\sum_{i=1}^{N} \left\| f'_i - {f_{gt}}_i \right\|_2^2
\end{equation}
where $b'_i = \left\{{b'_i}^f\right\}_{f=1}^{F}$ is the predicted probability distribution of different prediction steps for agent $i$, $N$ is the number of agents in the dataset, $\mathcal{L}_{cls}$ is the cross entropy loss function, $\mathcal{L}_{reg}$ is the regression loss, and $b_i$ is the one-hot encoded ground truth with ${b_i^j=\mathbb{\mathrm{1}}[j=f_i]}$. The APM is trained to predict the optimal prediction step based on the input context.\par
In summary, the loss function for trainging the APM is:
\begin{equation}
    \mathcal{L}_{APM} = \mathcal{L}_{cls} + \mathcal{L}_{reg}
\end{equation}
\subsubsection{Inference stage}
APM is applied to the training and inference process of FSN as a plug-and-play module. APM takes the encoded latent features of the target agent and its context as input and outputs the predicted optimal prediction step:
\begin{equation}
    f_i = \phi_{step}(e_{i,k})
\end{equation}
where $f_i$ is the predicted optimal prediction step for agent $i$. The APM can be integrated into any trajectory prediction model, allowing it to dynamically adjust the prediction steps based on the contextual information and environmental conditions. This adaptability is crucial for ensuring that the model can effectively handle varying prediction requirements in real-world scenarios.\par
\subsection{Dynamic Decoder}\label{method:dd}
Traditional trajectory prediction models typically use one or two layers Multilayer Perceptron (MLP) as the decoder to decode the latent embedding into future trajectory with fixed steps, as shown in eq.~\ref{eq:fixed_output}. This leads to a limitation in flexibility, as the model can only align weights to a specific output steps. To address this, we propose Dynamic Decoder (DD) that can handle varying prediction lengths, allowing the model to output sequences with different step lengths during inference.\par
Specifically, each time step of prediction matches the decoder with the corresponding step length. During training, the input $e_i$ and its output steps $f_i$, which is from the APM, are processed by their respective sub-networks $\left\{\phi_{D^{f}}\right\}_{f=5}^F$:
\begin{equation}
    \mu_{i,k}^{f_i}, b_{i,k}^{f_i} = \phi_{DD^{f_i}}(e_{i,k}, f_i)
\end{equation}
where $\mu_{i,k}^{f_i}$ and $b_{i,k}^{f_i}$ are the predicted future trajectory and probability of agent $i$ mode $k$ with output steps $f_i$, $\phi_{DD^{f_i}}$ is the decoder function for output steps $f_i$. During inference, the sub-network matching the predicted output steps $f'_i$ is exclusively activated.\par
Additionally, the input $e_{i,k}$ with different movement information may lead to different fit, which has potential drawbacks. To address this, we employ KL divergence\cite{li2024lakd} to distill knowledge from the "lower" score sequences to the "higher" ones:
\begin{equation}
    \mathcal{L}_{KL} = KL(V_{h}, V_{l})
\end{equation}
where $V_{h}$ and $V_{l}$ are the latent features of the "higher" and "lower" score sequences respectively.

\subsection{Training Objective}
In addition to the pre-trained APM, we also need to train the entire FlexiSteps Network. Following\cite{zhou2022hivt} and HPNet\cite{tang2024hpnet}, we also adopt the negative log-likelihood in HiVT and Huber in HPNet as the regression loss $\mathcal{L}_{reg}$. We also use the cross-entropy loss as the classification loss $\mathcal{L}_{cls}$ to optimize the model. Finally, the total loss function can be expressed as:
\begin{equation}
    \mathcal{L} = \mathcal{L}_{reg} + \mathcal{L}_{cls} + \lambda\mathcal{L}_{KL}
\end{equation}
where $\lambda$ is a hyperparameter to balance the contribution of KL loss function. The training process involves optimizing the model parameters to minimize the total loss, allowing the FlexiSteps Network to effectively learn from the data and adapt to varying prediction requirements.

\begin{figure*}[ht]
\centering
\subfloat[5 timesteps]{\includegraphics[width=.15\textwidth]{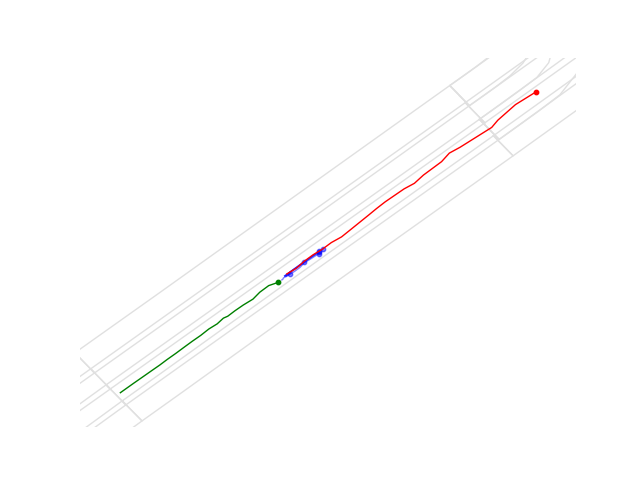}}\hspace{2pt}
\subfloat[10 timesteps]{\includegraphics[width=.15\textwidth]{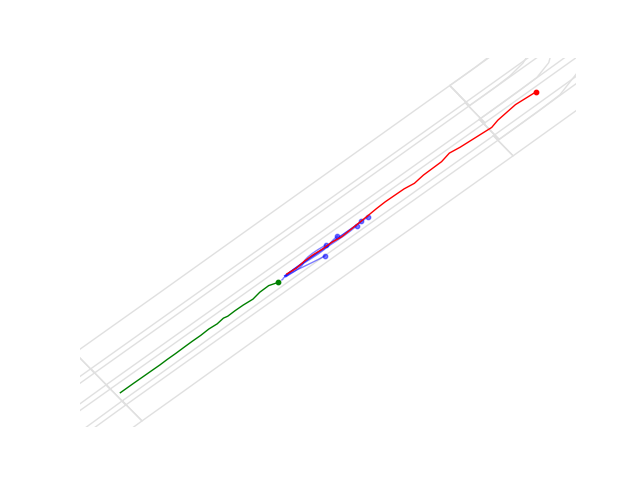}}\hspace{2pt}
\subfloat[15 timesteps]{\includegraphics[width=.15\textwidth]{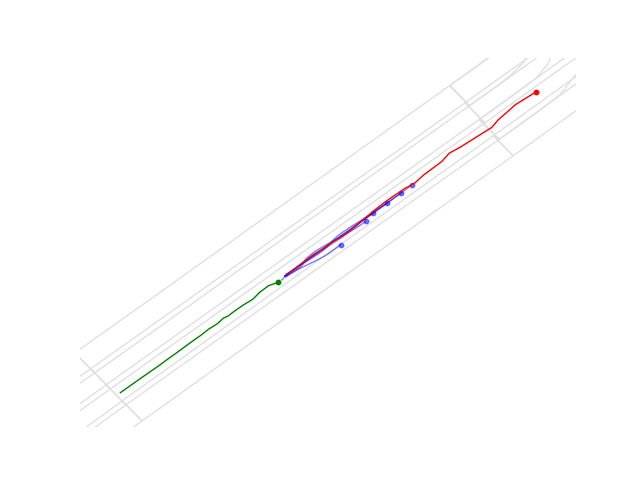}}\hspace{2pt}
\subfloat[20 timesteps]{\includegraphics[width=.15\textwidth]{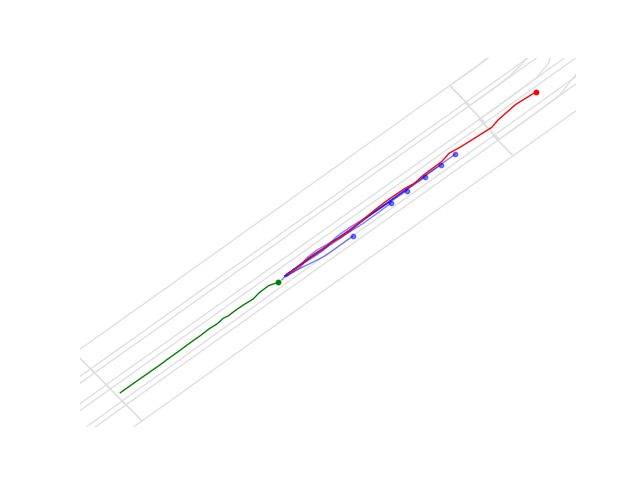}}\hspace{2pt}
\subfloat[25 timesteps]{\includegraphics[width=.15\textwidth]{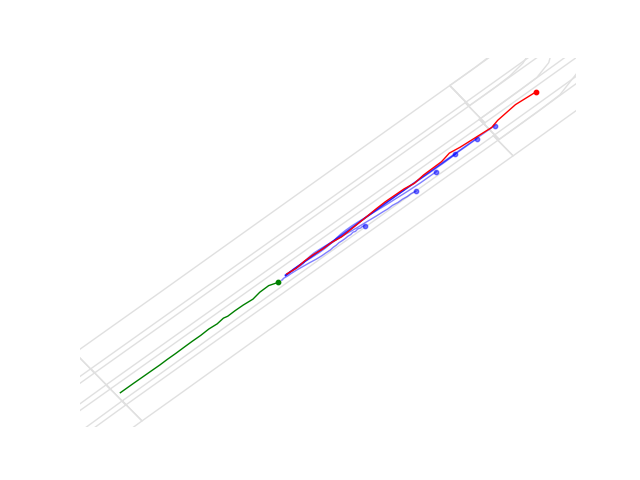}}\hspace{2pt}
\subfloat[30 timesteps]{\includegraphics[width=.15\textwidth]{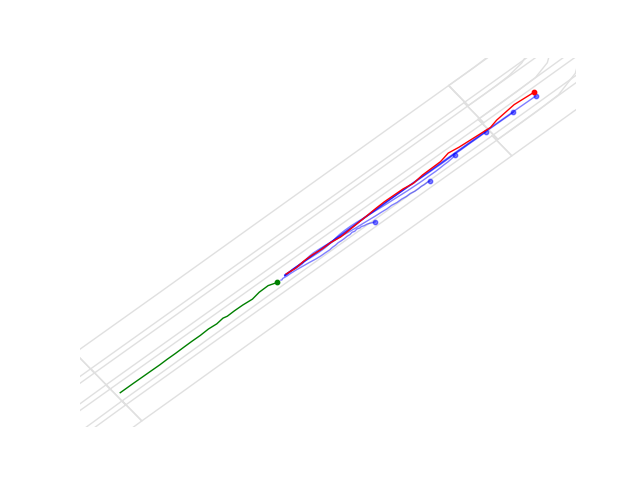}}\hspace{2pt}
\caption{The prediction results of HiVT with different prediction steps from Intercepted Results(IR). The blue line is the prediction trajectories, the red line is the ground truth trajectory, and the green line is the historical trajectory.}
\label{ir_result}
\end{figure*}

\begin{figure*}[ht]
\centering
\subfloat[5 timesteps]{\includegraphics[width=.15\textwidth]{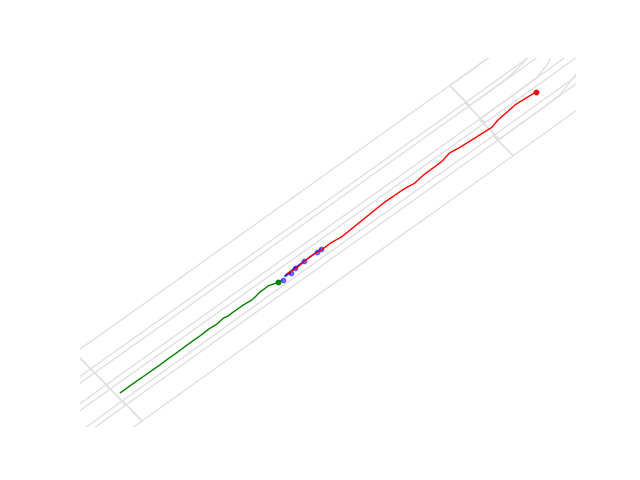}}\hspace{2pt}
\subfloat[10 timesteps]{\includegraphics[width=.15\textwidth]{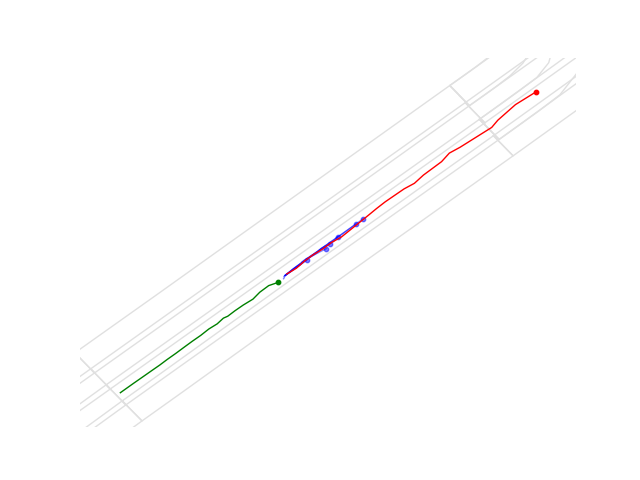}}\hspace{2pt}
\subfloat[15 timesteps]{\includegraphics[width=.15\textwidth]{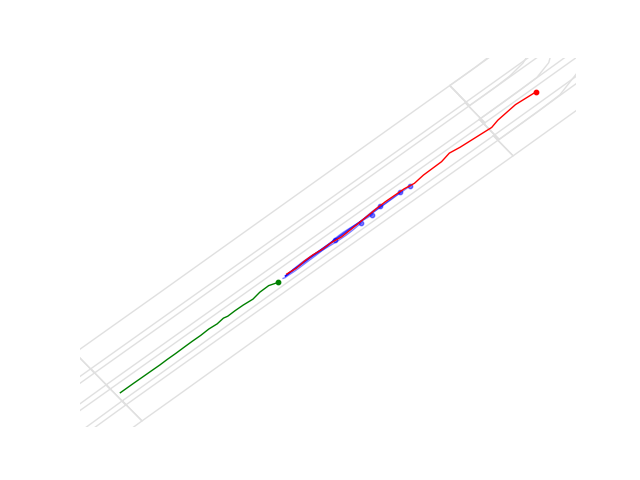}}\hspace{2pt}
\subfloat[20 timesteps]{\includegraphics[width=.15\textwidth]{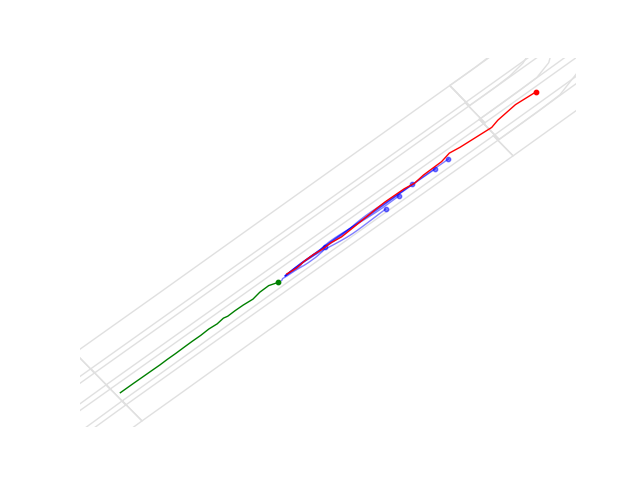}}\hspace{2pt}
\subfloat[25 timesteps]{\includegraphics[width=.15\textwidth]{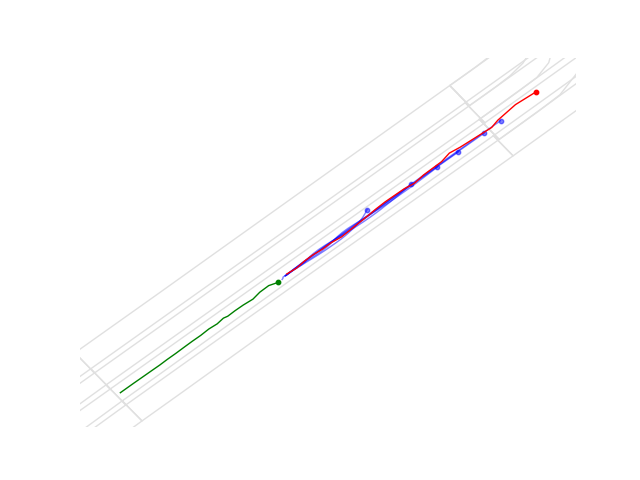}}\hspace{2pt}
\subfloat[30 timesteps]{\includegraphics[width=.15\textwidth]{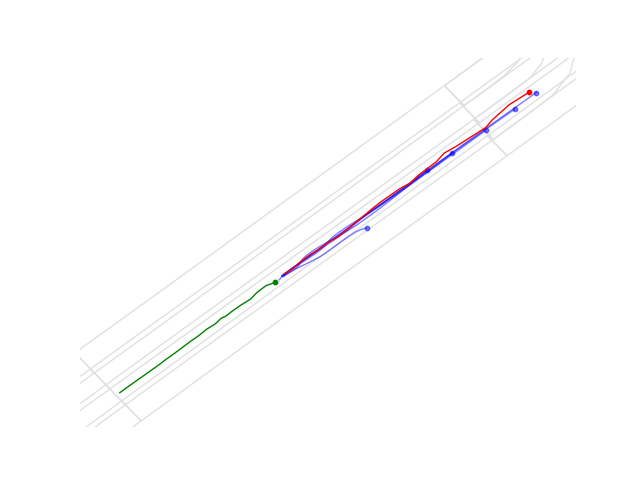}}\hspace{2pt}
\subfloat[flexible prediction]{\includegraphics[width=.15\textwidth]{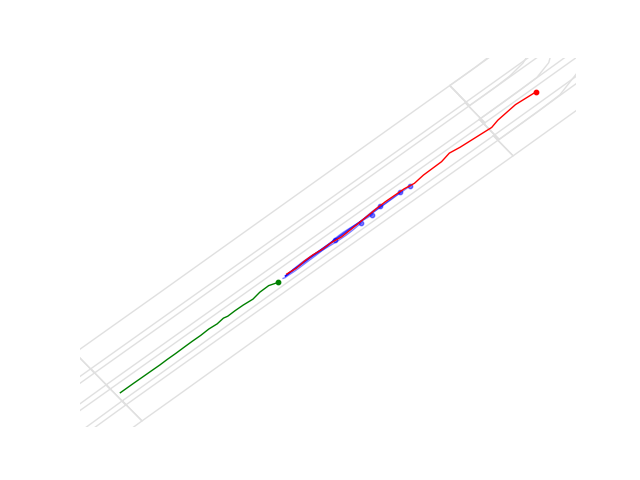}}
\caption{The prediction results of HiVT trained with fixed prediction steps. The blue line is the prediction trajectories, the red line is the ground truth trajectory, and the green line is the historical trajectory. The last subfigure is the prediction results of our FlexiSteps Network (FSN).}
\label{fixed_flexi_result}
\end{figure*}

\section{Expreiments}
\subsection{Settings}
\subsubsection{Baselines}
Our FlexiSteps Network (FSN) is a plug-and-play model that can be integrated into deep learning-based trajectory prediction models. We have chosen HiVT\cite{zhou2022hivt} and HPNet\cite{tang2024hpnet}, a typical method and a state-of-the-art open source method, as our baseline models. Due to limited computing resources,for the HPNet model, we only use the propose stage among the propose prediction and refine prediction stage. For more comprehensive evaluation, we include 2 additional baselines: (1) \textbf{Isolated Training} (IT): Train the baseline model with different fixed output steps. (2) \textbf{Intercepted Result} (IR): The model just trained once with the longest prediction steps and the prediction results are intercepted to different steps.
\subsubsection{Datasets}
\textbf{Argoverse} \cite{chang2019argoverse} is a large-scale dataset for autonomous driving, containing diverse scenarios and complex interactions between agents. It includes high-definition maps and rich contextual information, making it suitable for evaluating trajectory prediction models. The dataset is divided into training, validation, and test sets, with a total of 324,557 interesting vehicle trajectories. This rich dataset includes high-definition (HD) maps and recordings of sensor data, referred to as “log segments,” collected in two U.S. cities: Miami and Pittsburgh. These cities were chosen for their distinct urban driving challenges, including unique road geometries, local driving habits, and a variety of traffic conditions.\par
\textbf{INTERACTION} \cite{zhan2019interaction} is an extensive dataset developed for autonomous driving research, particularly focusing on behavioral aspects such as motion prediction, behavior analysis, and behavior cloning. It contains a diverse collection of natural movement patterns from various road users, including vehicles and pedestrians, captured across numerous highly interactive traffic scenarios from multiple countries. This dataset provides valuable information for studying complex agent interactions in different driving environments.

\begin{table*}[ht]
\centering
\renewcommand{\arraystretch}{1.5}
\resizebox{\textwidth}{!}{
\begin{tabular}{c|c|c|c|c|c|c|c}
\hline
\multirow{3}{*}{\textbf{Dataset}} & \multirow{3}{*}{\textbf{Method}} & \multicolumn{6}{c}{\textbf{Metrics}} \\ 
\cline{3-8}
 & & \multicolumn{6}{c}{\textbf{$\text{FDE}$}/\textbf{$\text{ADE}$}/\textbf{$\text{MR}$}} \\
\cline{3-8}
 & & $5\ timesteps$ & $10\ timesteps$ & $15\ timesteps$ & $20\ timesteps$ & $25\ timesteps$ & $30\ timesteps$ \\ \hline
\multirow{5}{*}{Argoverse 1} & HiVT-IT\cite{zhou2022hivt} & 0.1710/0.2203/0.0024 & 0.2782/0.2910/0.0052 & 0.4103/0.3638/0.0153 & 0.5714/0.4470/0.0331 & 0.7626/0.5442/0.0592 & 0.9701/0.6602/0.0923 \\
 & HiVT-IR\cite{zhou2022hivt} & 0.3635/0.2904/0.0107 & 0.5244/0.3753/0.0240 & 0.6701/0.4542/0.0433 & 0.8038/0.5283/0.0672 & 0.9219/0.5974/0.0915 & 0.9698/0.6611/0.0921 \\
 & HiVT-FLN\cite{xu2024adapting} & - & - & - & - & - & 1.0325/0.7026/0.1033 \\
 & HiVT-LaKD\cite{li2024lakd} & - & - & - & - & - & 0.9864/0.6807/0.0928 \\
 & HiVT-FSN(ours) & \textbf{0.1632}/\textbf{0.2121}/\textbf{0.0020} & \textbf{0.2678}/\textbf{0.2804}/\textbf{0.0045} & \textbf{0.3952}/\textbf{0.3501}/\textbf{0.0129} & \textbf{0.5503}/\textbf{0.4302}/\textbf{0.0287} & \textbf{0.7501}/\textbf{0.5393}/\textbf{0.0512} & \textbf{0.9602}/\textbf{0.6571}/\textbf{0.0904} \\ \hline
\hline
\multirow{3}{*}{Argoverse 1} & HPNet-IT\cite{tang2024hpnet}(need real data) & 0.1621-/0.2102-/0.0023- & 0.2990/0.2905/0.0045 & 0.4177/0.3717/\textbf{0.0111} & 0.5813/0.4576/0.0278 & 0.7685/0.5579/0.0501 & 0.9751/0.6790/0.0838 \\
 & HPNet-IR\cite{tang2024hpnet} & 0.3818/0.3037/0.0109 & 0.5491/0.3933/0.0238 & 0.6924/0.4747/0.0419 & 0.8232/0.5490/0.0657 & 0.9340/0.6169/0.0873 & 0.9741/0.6784/0.0829 \\
 & HPNet-FSN(ours)(need to reconsider) & \textbf{0.1583}/\textbf{0.2064}/\textbf{0.0019} & \textbf{0.2853}/\textbf{0.2862}/\textbf{0.0042} & \textbf{0.4001}/\textbf{0.3602}/\text{0.0118} & \textbf{0.5726}/\textbf{0.4403}/\textbf{0.0269} & \textbf{0.7574}/\textbf{0.5504}/\textbf{0.0487} & \textbf{0.9624}/\textbf{0.6678}/\textbf{0.0786} \\ \hline
 \hline
\multirow{3}{*}{INTERACTION} & HiVT-IT\cite{zhou2022hivt} & 0.0067/0.0069 & 0.0197/0.0153 & 0.0587/0.0245 & 0.1577/\textbf{0.0578} & 0.3058/0.1104 & 0.5963/0.2034 \\
 & HiVT-IR\cite{zhou2022hivt} & 0.0140/0.0087 & 0.0368/0.0179 & 0.0930/0.0347 & 0.1980/0.0644 & 0.3577/0.1096 & 0.5954/0.2057 \\
 & HiVT-FSN(ours) & \textbf{0.0067}/\textbf{0.0066} & \textbf{0.0172}/\textbf{0.0139} & \textbf{0.0496}/\textbf{0.0186} & \textbf{0.1496}/\text{0.0584} & \textbf{0.2874}/\textbf{0.1058} & \textbf{0.5829}/\textbf{0.1927} \\ \hline
 \hline
\multirow{3}{*}{INTERACTION} & HPNet-IT\cite{tang2024hpnet} & 0.0068/0.0069 & 0.0205/0.0146 & 0.0634/0.0280 & 0.1501/0.0549 & 0.3008/0.1053 & 0.5721/0.1761 \\
 & HPNet-IR\cite{tang2024hpnet} & 0.0148/0.0092 & 0.0388/0.0188 & 0.0962/0.0364 & 0.2029/0.0670 & 0.3625/0.1133 & 0.5754/0.1765 \\
 & HPNet-FSN(ours) & \textbf{0.0064}/\textbf{0.0066} & \textbf{0.0194}/\textbf{0.0122} & \textbf{0.0542}/\textbf{0.0201} & \textbf{0.1378}/\textbf{0.0468} & \textbf{0.2976}/\textbf{0.0936} & \textbf{0.5645}/\textbf{0.1647} \\ \hline
\end{tabular}
}
\caption{Performance comparison of FlexiSteps Network (FSN) with baselines on Argoverse and INTERACTION datasets. The best results are highlighted in bold. The "-" indicates that the method does not open source and the results are not available.}
\label{tab1}
\end{table*}

\subsubsection{Metrics}
Although we have metioned the limitations of mainstream metrics such as ADE and FDE, we still use them because we need to compare our FSN with traditional methods to verify the performance of our method.\par
Specifically, for Argoverse dataset, we use minimum ADE (minADE), minimum FDE (minFDE) and Miss Rate (MR) to measure the accuracy and robustness of the model. The minimum ADE is the minimum average displacement error between the predicted trajectory and the ground truth trajectory, while the minimum FDE is the minimum final displacement error. The Miss Rate is the percentage of predicted trajectories that do not match any ground truth trajectory within a certain threshold.\par
For INTERACTION dataset, we employ minJointADE, minJointFDE and Cross Collision Rate (CCR) to evaluate the performance of joint trajectory prediction. The minJointADE is the minimum average displacement error between the predicted joint trajectory and the ground truth joint trajectory, while the minJointFDE is the minimum final displacement error. The Cross Collision Rate (CCR) is the percentage of predicted joint trajectories that collide with each other, indicating the model's ability to handle interactions between agents.\par
\subsubsection{Implementation Details}
For the HiVT baseline, we train our APM and FSN for 64 epochs with a batch size of 32 on 1 NVIDIA A6000 GPU, using the AdamW optimizer with a learning rate of $5\times 10^{-4}$ and weight decay of $1\times 10^{-4}$. For the HPNet baseline, we train our APM and FSN for 64 epochs with a batch size of 4 on 1 NVIDIA A6000 GPU, using the AdamW optimizer with a learning rate of $5\times 10^{-4}$ and weight decay of $1\times 10^{-4}$.\par
The hyperparameter $\lambda$ is set to 0.5 for both baselines. The training process involves optimizing the model parameters to minimize the total loss, allowing the FlexiSteps Network to effectively learn from the data and adapt to varying prediction requirements.

\subsection{Main Results}
\subsubsection{Argoverse}
As shown in Table~\ref{tab1}, our FlexiSteps Network (FSN) outperforms the baselines on the Argoverse dataset across all prediction steps. Specifically, FSN achieves a minimum ADE of 0.2121 and a minimum FDE of 0.1632 at 5 timesteps, which is significantly better than the IT and IR baselines. The results demonstrate that FSN effectively leverages the pre-trained Adaptive Prediction Module (APM) and Dynamic Decoder (DD) to adaptively adjust the prediction steps based on contextual information, leading to improved accuracy and robustness in trajectory prediction.\par
In addition, we also compare our FSN with the state-of-the-art dynamic prediction methods FLN and LaKD. The results show that FSN achieves better performance than FLN and LaKD, demonstrating the effectiveness of our approach in handling varying prediction lengths. The minimum ADE and FDE of FSN at 30 timesteps are 0.6571 and 0.9602, respectively, which are significantly lower than those of FLN and LaKD. This indicates that FSN can effectively adapt to different prediction requirements and achieve better performance in trajectory prediction tasks.\par
\subsubsection{INTERACTION}
As shown in Table~\ref{tab1}, our FlexiSteps Network (FSN) also outperforms the baselines on the INTERACTION dataset across all prediction steps. Specifically, FSN achieves a minimum Joint ADE of 0.0066 and a minimum Joint FDE of 0.0067 at 5 timesteps, which is significantly better than the IT and IR baselines. The results demonstrate that FSN effectively leverages the pre-trained Adaptive Prediction Module (APM) and Dynamic Decoder (DD) to adaptively adjust the prediction steps based on contextual information, leading to improved accuracy and robustness in joint trajectory prediction.\par

\subsection{Ablation Study}
To evaluate the effectiveness of each component in our FlexiSteps Network (FSN), we conduct an ablation study on the Argoverse validation dataset with 30 prediction steps. The results are presented in Table~\ref{tab2}. We analyze the impact of the Dynamic Decoder (DD), and scoring mechanism on the overall performance of FSN. Specifically, we compare the performance of Dynamic Decoder (DD) with and without the KL divergence loss, and the performance of the scoring mechanism with and without the Fréchet distance, we employ ADE and FDE instead of Fréchet distance.\par

\begin{table*}[t]
    \centering
    \begin{tabular}{llllllll}
        \hline
\multicolumn{3}{c}{Score} & \multicolumn{2}{c}{Dynamic Decoder} & \multirow{2}{*}{FDE} & \multirow{2}{*}{ADE} & \multirow{2}{*}{MR} \\ \cline{1-5}
    Fréchet  &   FDE   &   ADE   &     w/o KL       &     w/ KL     &                   &                   &                   \\ \hline
      &   \checkmark    &      &           &          &          0.9705         &        0.6602          &         0.0922          \\ 
      &       &   \checkmark   &           &          &         0.9727          &         0.6594         &         0.0923          \\ 
    \checkmark  &       &      &     \checkmark     &          &         0.9683          &        0.6647           &         0.0921          \\ 
    \checkmark  &       &      &           &     \checkmark     &         \textbf{0.9602}          &         \textbf{0.6571}          &         \textbf{0.0904} \\         \hline
\end{tabular}
    \caption{Ablation study of FlexiSteps Network (FSN) with HiVT as our backbone on Argoverse validation dataset. The results show the impact of different components on the overall performance of FSN. The best results are highlighted in bold.}
    \label{tab2}
\end{table*}

The results show that our score mechanism with Fréchet distance achieves the best performance, with a minimum FDE of 0.9602 and a minimum ADE of 0.6571, significantly outperforming the score mechanism with ADE and FDE. This indicates that the Fréchet distance effectively captures the spatial and temporal relationships in trajectory predictions, leading to improved accuracy. And the Dynamic Decoder (DD) with KL divergence loss also improves the performance, achieving a minimum FDE of 0.9683 and a minimum ADE of 0.6647, compared to the DD without KL divergence loss. This demonstrates that the KL divergence loss helps to distill knowledge from lower score sequences to higher ones, enhancing the model's ability to adapt to varying prediction lengths.\par

\section{Conclusion}
In this paper, we propose the FlexiSteps Network (FSN), a novel framework for dynamic trajectory prediction that adapts the prediction output steps based on contextual cues and environmental conditions. FSN consists of three main components: a pre-trained Adaptive Prediction Module (APM), a Dynamic Decoder (DD), and a scoring mechanism that incorporates the Fréchet distance to evaluate trajectory predictions. The APM is trained to predict the optimal prediction step based on the encoded latent features of the target agent and its context, while the DD handles varying prediction lengths. The scoring mechanism evaluates the quality of trajectory predictions by combining the Fréchet distance with the prediction length. Our experimental results demonstrate that FSN outperforms state-of-the-art methods on both Argoverse and INTERACTION datasets, achieving better accuracy and robustness in trajectory prediction tasks.\par
While our FlexiSteps Network (FSN) shows promising results in dynamic trajectory prediction, there are still several limitations and potential areas for future work. One limitation is the reliance on the pre-trained Adaptive Prediction Module (APM), which may not generalize well to unseen scenarios or environments. Future work could explore methods to enhance the adaptability of APM, such as incorporating online learning or transfer learning techniques. Additionally, the scoring mechanism based on Fréchet distance may not capture all aspects of trajectory quality, and further research could investigate alternative metrics or hybrid approaches to improve evaluation accuracy. Finally, while FSN demonstrates flexibility in handling varying prediction lengths, it may still struggle with highly dynamic or unpredictable environments. Future research could focus on enhancing the model's robustness to such scenarios, potentially through the integration of reinforcement learning or other adaptive techniques.

\section{Data Availability Statement}
The data underlying this article are available in \textit{Argoverse 1 Motion Forcasting Dataset} and \textit{INTERnational, Adversarial and Cooperative moTION Dataset}, at \url{https://dx.doi.org/10.48550/arXiv.1911.02620} and \url{https://dx.doi.org/10.48550/arXiv.1910.03088}.

\bibliographystyle{ieeetr}
\bibliography{IEEEabrv,mylib}

\begin{thebibliography}{10}

\bibitem{chai2019multipath}
Y.~Chai, B.~Sapp, M.~Bansal, and D.~Anguelov, ``Multipath: Multiple probabilistic anchor trajectory hypotheses for behavior prediction,'' {\em arXiv preprint arXiv:1910.05449}, 2019.

\bibitem{varadarajan2022multipath++}
B.~Varadarajan, A.~Hefny, A.~Srivastava, K.~S. Refaat, N.~Nayakanti, A.~Cornman, K.~Chen, B.~Douillard, C.~P. Lam, D.~Anguelov, {\em et~al.}, ``Multipath++: Efficient information fusion and trajectory aggregation for behavior prediction,'' in {\em 2022 International Conference on Robotics and Automation (ICRA)}, pp.~7814--7821, IEEE, 2022.

\bibitem{zhou2022hivt}
Z.~Zhou, L.~Ye, J.~Wang, K.~Wu, and K.~Lu, ``Hivt: Hierarchical vector transformer for multi-agent motion prediction,'' in {\em Proceedings of the IEEE/CVF Conference on Computer Vision and Pattern Recognition}, pp.~8823--8833, 2022.

\bibitem{zhou2023query}
Z.~Zhou, J.~Wang, Y.-H. Li, and Y.-K. Huang, ``Query-centric trajectory prediction,'' in {\em Proceedings of the IEEE/CVF Conference on Computer Vision and Pattern Recognition}, pp.~17863--17873, 2023.

\bibitem{zhou2024smartrefine}
Y.~Zhou, H.~Shao, L.~Wang, S.~L. Waslander, H.~Li, and Y.~Liu, ``Smartrefine: A scenario-adaptive refinement framework for efficient motion prediction,'' in {\em Proceedings of the IEEE/CVF Conference on Computer Vision and Pattern Recognition}, pp.~15281--15290, 2024.

\bibitem{tang2024hpnet}
X.~Tang, M.~Kan, S.~Shan, Z.~Ji, J.~Bai, and X.~Chen, ``Hpnet: Dynamic trajectory forecasting with historical prediction attention,'' in {\em Proceedings of the IEEE/CVF Conference on Computer Vision and Pattern Recognition}, pp.~15261--15270, 2024.

\bibitem{wang2025dynamics}
C.~Wang, H.~Liao, K.~Zhu, G.~Zhang, and Z.~Li, ``A dynamics-enhanced learning model for multi-horizon trajectory prediction in autonomous vehicles,'' {\em Information Fusion}, vol.~118, p.~102924, 2025.

\bibitem{climent2025learning}
J.~C. Climent~Pardo, ``Learning isometric embeddings of road networks using multidimensional scaling,'' {\em arXiv e-prints}, pp.~arXiv--2504, 2025.

\bibitem{liu2025gamdtp}
Y.~Liu, H.~Niu, and J.~Zhu, ``Gamdtp: Dynamic trajectory prediction with graph attention mamba network,'' {\em arXiv preprint arXiv:2504.04862}, 2025.

\bibitem{zhu2025dyttp}
J.~Zhu and H.~Niu, ``Dyttp: Trajectory prediction with normalization-free transformers,'' {\em arXiv preprint arXiv:2504.05356}, 2025.

\bibitem{xu2024adapting}
Y.~Xu and Y.~Fu, ``Adapting to length shift: Flexilength network for trajectory prediction,'' in {\em Proceedings of the IEEE/CVF Conference on Computer Vision and Pattern Recognition}, pp.~15226--15237, 2024.

\bibitem{li2024lakd}
Y.~Li, C.~Li, R.~Lv, R.~Li, Y.~Yuan, and G.~Wang, ``Lakd: Length-agnostic knowledge distillation for trajectory prediction with any length observations,'' {\em Advances in Neural Information Processing Systems}, vol.~37, pp.~28720--28744, 2024.

\bibitem{he2024frechet}
S.~He and C.~Wu, ``A fr{\'e}chet distance-hierarchical density-based spatial clustering of applications with noise for ship trajectory clustering and route recognition,'' in {\em 2024 International Conference on Distributed Systems, Computer Networks and Cybersecurity (ICDSCNC)}, pp.~1--5, IEEE, 2024.

\bibitem{kuo2022trajectory}
Y.-L. Kuo, X.~Huang, A.~Barbu, S.~G. McGill, B.~Katz, J.~J. Leonard, and G.~Rosman, ``Trajectory prediction with linguistic representations,'' in {\em 2022 International Conference on Robotics and Automation (ICRA)}, pp.~2868--2875, IEEE, 2022.

\bibitem{eiter1994computing}
T.~Eiter and H.~Mannila, ``Computing discrete fr{\'e}chet distance,'' 1994.

\bibitem{10939571}
S.~He and C.~Wu, ``A fréchet distance -hierarchical density-based spatial clustering of applications with noise for ship trajectory clustering and route recognition,'' in {\em 2024 International Conference on Distributed Systems, Computer Networks and Cybersecurity (ICDSCNC)}, pp.~1--5, 2024.

\bibitem{chang2019argoverse}
M.-F. Chang, J.~Lambert, P.~Sangkloy, J.~Singh, S.~Bak, A.~Hartnett, D.~Wang, P.~Carr, S.~Lucey, D.~Ramanan, {\em et~al.}, ``Argoverse: 3d tracking and forecasting with rich maps,'' in {\em Proceedings of the IEEE/CVF conference on computer vision and pattern recognition}, pp.~8748--8757, 2019.

\bibitem{zhan2019interaction}
W.~Zhan, L.~Sun, D.~Wang, H.~Shi, A.~Clausse, M.~Naumann, J.~Kummerle, H.~Konigshof, C.~Stiller, A.~de~La~Fortelle, {\em et~al.}, ``Interaction dataset: An international, adversarial and cooperative motion dataset in interactive driving scenarios with semantic maps,'' {\em arXiv preprint arXiv:1910.03088}, 2019.

\bibitem{liang2020learning}
M.~Liang, B.~Yang, R.~Hu, Y.~Chen, R.~Liao, S.~Feng, and R.~Urtasun, ``Learning lane graph representations for motion forecasting,'' in {\em Computer Vision--ECCV 2020: 16th European Conference, Glasgow, UK, August 23--28, 2020, Proceedings, Part II 16}, pp.~541--556, Springer, 2020.

\bibitem{wang2020multi}
Y.~Wang, S.~Zhao, R.~Zhang, X.~Cheng, and L.~Yang, ``Multi-vehicle collaborative learning for trajectory prediction with spatio-temporal tensor fusion,'' {\em IEEE Transactions on Intelligent Transportation Systems}, vol.~23, no.~1, pp.~236--248, 2020.

\bibitem{xu2023mvhgn}
D.~Xu, X.~Shang, H.~Peng, and H.~Li, ``Mvhgn: Multi-view adaptive hierarchical spatial graph convolution network based trajectory prediction for heterogeneous traffic-agents,'' {\em IEEE Transactions on Intelligent Transportation Systems}, vol.~24, no.~6, pp.~6217--6226, 2023.

\bibitem{zhang2022ai}
K.~Zhang, L.~Zhao, C.~Dong, L.~Wu, and L.~Zheng, ``Ai-tp: Attention-based interaction-aware trajectory prediction for autonomous driving,'' {\em IEEE Transactions on Intelligent Vehicles}, vol.~8, no.~1, pp.~73--83, 2022.

\bibitem{sheng2022graph}
Z.~Sheng, Y.~Xu, S.~Xue, and D.~Li, ``Graph-based spatial-temporal convolutional network for vehicle trajectory prediction in autonomous driving,'' {\em IEEE Transactions on Intelligent Transportation Systems}, vol.~23, no.~10, pp.~17654--17665, 2022.

\bibitem{goodfellow2014generative}
I.~J. Goodfellow, J.~Pouget-Abadie, M.~Mirza, B.~Xu, D.~Warde-Farley, S.~Ozair, A.~Courville, and Y.~Bengio, ``Generative adversarial nets,'' {\em Advances in neural information processing systems}, vol.~27, 2014.

\bibitem{gupta2018social}
A.~Gupta, J.~Johnson, L.~Fei-Fei, S.~Savarese, and A.~Alahi, ``Social gan: Socially acceptable trajectories with generative adversarial networks,'' in {\em Proceedings of the IEEE conference on computer vision and pattern recognition}, pp.~2255--2264, 2018.

\bibitem{azadani2023stag}
M.~N. Azadani and A.~Boukerche, ``Stag: A novel interaction-aware path prediction method based on spatio-temporal attention graphs for connected automated vehicles,'' {\em Ad Hoc Networks}, vol.~138, p.~103021, 2023.

\bibitem{gu2021densetnt}
J.~Gu, C.~Sun, and H.~Zhao, ``Densetnt: End-to-end trajectory prediction from dense goal sets,'' in {\em Proceedings of the IEEE/CVF International Conference on Computer Vision}, pp.~15303--15312, 2021.

\bibitem{ngiam2021scene}
J.~Ngiam, B.~Caine, V.~Vasudevan, Z.~Zhang, H.-T.~L. Chiang, J.~Ling, R.~Roelofs, A.~Bewley, C.~Liu, A.~Venugopal, {\em et~al.}, ``Scene transformer: A unified architecture for predicting multiple agent trajectories,'' {\em arXiv preprint arXiv:2106.08417}, 2021.

\bibitem{mohamed2020social}
A.~Mohamed, K.~Qian, M.~Elhoseiny, and C.~Claudel, ``Social-stgcnn: A social spatio-temporal graph convolutional neural network for human trajectory prediction,'' in {\em Proceedings of the IEEE/CVF conference on computer vision and pattern recognition}, pp.~14424--14432, 2020.

\bibitem{gao2023dual}
K.~Gao, X.~Li, B.~Chen, L.~Hu, J.~Liu, R.~Du, and Y.~Li, ``Dual transformer based prediction for lane change intentions and trajectories in mixed traffic environment,'' {\em IEEE Transactions on Intelligent Transportation Systems}, vol.~24, no.~6, pp.~6203--6216, 2023.

\bibitem{zhou2022spatiotemporal}
X.~Zhou, W.~Zhao, A.~Wang, C.~Wang, and S.~Zheng, ``Spatiotemporal attention-based pedestrian trajectory prediction considering traffic-actor interaction,'' {\em IEEE Transactions on Vehicular Technology}, vol.~72, no.~1, pp.~297--311, 2022.

\bibitem{hu2022trajectory}
H.~Hu, Q.~Wang, M.~Cheng, and Z.~Gao, ``Trajectory prediction neural network and model interpretation based on temporal pattern attention,'' {\em IEEE Transactions on Intelligent Transportation Systems}, vol.~24, no.~3, pp.~2746--2759, 2022.

\bibitem{hou2022structural}
L.~Hou, S.~E. Li, B.~Yang, Z.~Wang, and K.~Nakano, ``Structural transformer improves speed-accuracy trade-off in interactive trajectory prediction of multiple surrounding vehicles,'' {\em IEEE Transactions on Intelligent Transportation Systems}, vol.~23, no.~12, pp.~24778--24790, 2022.

\bibitem{karle2024self}
P.~Karle, L.~Furtner, and M.~Lienkamp, ``Self-evaluation of trajectory predictors for autonomous driving,'' {\em Electronics}, vol.~13, no.~5, p.~946, 2024.

\bibitem{qiao2014self}
S.~Qiao, D.~Shen, X.~Wang, N.~Han, and W.~Zhu, ``A self-adaptive parameter selection trajectory prediction approach via hidden markov models,'' {\em IEEE Transactions on Intelligent Transportation Systems}, vol.~16, no.~1, pp.~284--296, 2014.

\bibitem{jiang2022vehicle}
Y.~Jiang, B.~Zhu, S.~Yang, J.~Zhao, and W.~Deng, ``Vehicle trajectory prediction considering driver uncertainty and vehicle dynamics based on dynamic bayesian network,'' {\em IEEE Transactions on Systems, Man, and Cybernetics: Systems}, vol.~53, no.~2, pp.~689--703, 2022.

\bibitem{zyner2019naturalistic}
A.~Zyner, S.~Worrall, and E.~Nebot, ``Naturalistic driver intention and path prediction using recurrent neural networks,'' {\em IEEE transactions on intelligent transportation systems}, vol.~21, no.~4, pp.~1584--1594, 2019.

\bibitem{chen2022intention}
X.~Chen, H.~Zhang, F.~Zhao, Y.~Hu, C.~Tan, and J.~Yang, ``Intention-aware vehicle trajectory prediction based on spatial-temporal dynamic attention network for internet of vehicles,'' {\em IEEE Transactions on Intelligent Transportation Systems}, vol.~23, no.~10, pp.~19471--19483, 2022.

\bibitem{zhang2022trajectory}
K.~Zhang, X.~Feng, L.~Wu, and Z.~He, ``Trajectory prediction for autonomous driving using spatial-temporal graph attention transformer,'' {\em IEEE Transactions on Intelligent Transportation Systems}, vol.~23, no.~11, pp.~22343--22353, 2022.

\bibitem{zhao2021tnt}
H.~Zhao, J.~Gao, T.~Lan, C.~Sun, B.~Sapp, B.~Varadarajan, Y.~Shen, Y.~Shen, Y.~Chai, C.~Schmid, {\em et~al.}, ``Tnt: Target-driven trajectory prediction,'' in {\em Conference on robot learning}, pp.~895--904, PMLR, 2021.

\bibitem{song2025don}
Z.~Song, C.~Jia, L.~Liu, H.~Pan, Y.~Zhang, J.~Wang, X.~Zhang, S.~Xu, L.~Yang, and Y.~Luo, ``Don't shake the wheel: Momentum-aware planning in end-to-end autonomous driving,'' in {\em Proceedings of the Computer Vision and Pattern Recognition Conference}, pp.~22432--22441, 2025.

\bibitem{huttenlocher1993comparing}
D.~P. Huttenlocher, G.~A. Klanderman, and W.~J. Rucklidge, ``Comparing images using the hausdorff distance,'' {\em IEEE Transactions on pattern analysis and machine intelligence}, vol.~15, no.~9, pp.~850--863, 1993.

\bibitem{dubuisson1994modified}
M.-P. Dubuisson and A.~K. Jain, ``A modified hausdorff distance for object matching,'' in {\em Proceedings of 12th international conference on pattern recognition}, vol.~1, pp.~566--568, IEEE, 1994.

\bibitem{belogay1997calculating}
E.~Belogay, C.~Cabrelli, U.~Molter, and R.~Shonkwiler, ``Calculating the hausdorff distance between curves,'' {\em Information Processing Letters}, vol.~64, no.~1, 1997.

\bibitem{shahbaz2013applied}
K.~Shahbaz, ``Applied similarity problems using fr{\'e}chet distance,'' {\em arXiv preprint arXiv:1307.6628}, 2013.

\bibitem{zhao20183d}
J.-L. Zhao, Z.-K. Wu, Z.-K. Pan, F.-Q. Duan, J.-H. Li, Z.-H. Lv, K.~Wang, and Y.-C. Chen, ``3d face similarity measure by fr{\'e}chet distances of geodesics,'' {\em Journal of Computer Science and Technology}, vol.~33, pp.~207--222, 2018.

\bibitem{bhattacharyya2018long}
A.~Bhattacharyya, M.~Fritz, and B.~Schiele, ``Long-term on-board prediction of people in traffic scenes under uncertainty,'' in {\em Proceedings of the IEEE conference on computer vision and pattern recognition}, pp.~4194--4202, 2018.

\bibitem{alt1995computing}
H.~Alt and M.~Godau, ``Computing the fr{\'e}chet distance between two polygonal curves,'' {\em International Journal of Computational Geometry \& Applications}, vol.~5, no.~01n02, pp.~75--91, 1995.

\bibitem{10.1145/3474717.3483949}
K.~Takeuchi, M.~Imaizumi, S.~Kanda, Y.~Tabei, K.~Fujii, K.~Yoda, M.~Ishihata, and T.~Maekawa, ``Fr\'{e}chet kernel for trajectory data analysis,'' in {\em Proceedings of the 29th International Conference on Advances in Geographic Information Systems}, SIGSPATIAL '21, (New York, NY, USA), p.~221–224, Association for Computing Machinery, 2021.

\bibitem{gupta2020robust}
D.~Gupta, B.~B. Hazarika, and M.~Berlin, ``Robust regularized extreme learning machine with asymmetric huber loss function,'' {\em Neural Computing and Applications}, vol.~32, no.~16, pp.~12971--12998, 2020.

\bibitem{meyer2021alternative}
G.~P. Meyer, ``An alternative probabilistic interpretation of the huber loss,'' in {\em Proceedings of the ieee/cvf conference on computer vision and pattern recognition}, pp.~5261--5269, 2021.

\end{thebibliography}
\end{document}